\documentclass[10pt,twocolumn,letterpaper]{article}

\usepackage{iccv}
\usepackage{times}
\usepackage{epsfig}
\usepackage{graphicx}
\usepackage{amsmath}
\usepackage{amssymb}

\usepackage{threeparttable}

\renewcommand{\mathbf}{\boldsymbol}
\newcommand{\FSTCN}{\textrm{F}_{\textrm{ST}}\textrm{CN}}
\iccvfinalcopy % *** Uncomment this line for the final submission

\def\iccvPaperID{1963} % *** Enter the ICCV Paper ID here

\ificcvfinal\fi
\begin{document}

\language0
\lefthyphenmin=2
\righthyphenmin=3

%%%%%%%%% TITLE
\title{Human Action Recognition using Factorized Spatio-Temporal \\ Convolutional Networks}

\author{Lin Sun \dag\S, Kui Jia $\sharp$, Dit-Yan Yeung \ddag\dag, Bertram E. Shi \dag \\
\small
\dag~Department of Electronic and Computer Engineering, Hong Kong University of Science and Technology\\
\small
\ddag~Department of Computer Science and Engineering, Hong Kong University of Science and Technology\\
\small
$\sharp$ Faculty of Science and Technology, University of Macau\\
\small
\S~Lenovo Corporate Research Hong Kong Branch\\
\small
{lsunece@ust.hk, kuijia@gmail.com, dyyeung@cse.ust.hk, eebert@ust.hk}}

\maketitle
\thispagestyle{empty}

%%%%%%%%% ABSTRACT
\begin{abstract}
{Human actions in video sequences are three-dimensional~(3D) spatio-temporal signals characterizing both the visual appearance and motion dynamics of the involved humans and objects. Inspired by the success of convolutional neural networks~(CNN) for image classification, recent attempts have been made to learn 3D CNNs for recognizing human actions in videos. However, partly due to the high complexity of training 3D convolution kernels and the need for large quantities of training videos, only limited success has been reported. This has triggered us to investigate in this paper a new deep architecture which can handle 3D signals more effectively. Specifically, we propose \emph{factorized spatio-temporal convolutional networks}~($\FSTCN$) that factorize the original 3D convolution kernel learning as a sequential process of learning 2D spatial kernels in the lower layers (called spatial convolutional layers), followed by learning 1D temporal kernels in the upper layers (called temporal convolutional layers). We introduce a novel transformation and permutation operator to make factorization in $\FSTCN$ possible. Moreover, to address the issue of sequence alignment, we propose an effective training and inference strategy based on sampling multiple video clips from a given action video sequence. We have tested $\FSTCN$ on two commonly used benchmark datasets (\mbox{UCF-101} and \mbox{HMDB-51}). Without using auxiliary training videos to boost the performance, $\FSTCN$ outperforms existing CNN based methods and achieves comparable performance with a recent method that benefits from using auxiliary training videos.}
\end{abstract}

%%%%%%%%% BODY TEXT
\section{Introduction} \label{Sec-Intro}
Human actions can be categorized by the visual appearance and motion dynamics of the involved humans and objects. The design of many popular human action recognition datasets \cite{UCF-101,HMDB-51,KTH} is based on this intrinsic property. To recognize human actions in video sequences, computer vision researchers have been developing better visual features to characterize the spatial appearance \cite{WeinlandB08,sift} and temporal motion \cite{HOF,BoW_2,vid_sfa}. Since video sequences can naturally be viewed as three-dimensional~(3D) spatio-temporal signals, many existing methods seek to develop different spatio-temporal features for representing spatially and temporally coupled action patterns \cite{hog3d,Wangs_evaluation,JiaYeung08,Irani_SpaceTimeAction}. Thus far, although these methods are robust against some real-world human action conditions, when applied to more realistic, complex human actions, their performance often degrades significantly due to the large intra-category variations within action categories and inter-category ambiguities between action categories. A number of factors can cause large intra-category variations. Some major ones include large variations in visual appearance and motion dynamics of the constituent humans and objects, arbitrary illumination and imaging conditions, self-occlusion, and cluttered background. To address these challenges, some methods~\cite{st_interest_points,improved-trajectories} extract trajectories of interest points from video sequences to characterize the salient spatial regions and their motion dynamics. However, in general, the challenge of recognizing complex human actions has not been well addressed.

Most of the above methods use handcrafted features and relatively simple classifiers. More recently, the end-to-end approach of learning features directly from raw observations using deep architectures shows great promise in many computer vision tasks, including object detection \cite{girshick2014rcnn}, semantic segmentation \cite{farabet2013pami} and so forth. Using massive training datasets, these deep architectures are able to learn a hierarchy of semantically related convolution filters (or kernels), giving highly discriminative models and hence better classification accuracy~\cite{Zeiler14}. In fact, even directly applying these image-based models to individual frames of the videos has shown promising action recognition performance \cite{vid_cnn,TwoStream}, because the learned features can better characterize the visual appearance in the spatial domain.

However, human actions in video sequences are 3D spatio-temporal signals. It is not surprising to expect that exploiting the temporal domain as well could further advance the state of the art. Some recent attempts have been made along this direction \cite{3dCNN,vid_cnn,TwoStream}. The 3D CNN model \cite{3dCNN} learns convolution kernels in both space and time based on a straightforward extension of the established 2D CNN deep architectures \cite{lecun,AlexNet} to the 3D spatio-temporal domain. The methods in \cite{vid_cnn} aim at learning long-range motion features by learning a hierarchy consisting of multiple layers of 3D spatio-temporal convolution kernels by early fusion, late fusion, or slow fusion. The two-stream CNN architecture \cite{TwoStream} learns motion patterns using an additional CNN which takes as input the optical flow computed from successive frames of video sequences. By using the optical flow to capture motion features, the two-stream \mbox{CNN} is less effective for characterizing long-range or ``slow'' motion patterns which may be more relevant to the semantic categorization of human actions \cite{vid_sfa,dlsfa}. Possibly due to the increased complexity and difficulty of training 3D kernels without sufficient training video data (as compared to massive image datasets~\cite{imagenet}), 3D CNN \cite{3dCNN} does not perform well even on the less challenging KTH dataset~\cite{KTH}. For the UCF-101 benchmark dataset~\cite{UCF-101}, we note that the results reported in~\cite{vid_cnn} are inferior to those obtained by two-stream \mbox{CNN}~\cite{TwoStream}. Indeed, spatio-temporal action patterns coupling the visual appearance and motion dynamics generally need an order of magnitude more training data than the 2D spatial counterparts. Moreover, existing methods often overlook the issue of sequence alignment in which actions of different speeds and accelerations have to be handled properly for human action recognition.

\begin{figure*}
  \centering
  \centerline{\includegraphics[width=15.5cm]{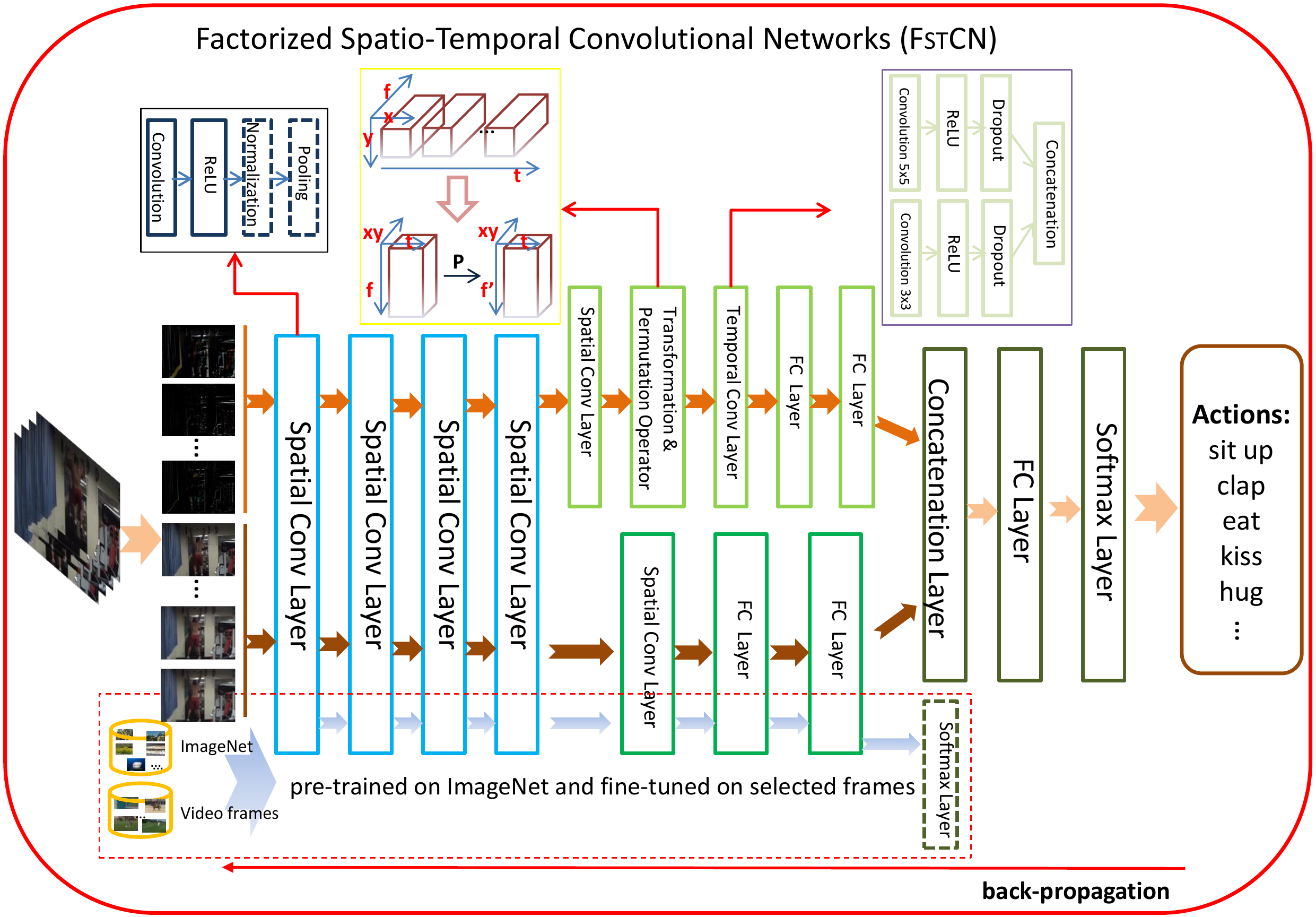}}
  \caption{Schematic diagram of $\FSTCN$ for action recognition. `Conv' is the short form for `convolutional' and `FC' for `fully connected'. Processing in the dashed boxes is optional.  More details are described in the paper.}\label{fig:FSTCN}\vspace{-1.5em}
\end{figure*}

The above analysis motivates us to consider alternative deep architectures which can handle 3D spatio-temporal signals more effectively.  To this end, we propose a new deep architecture called \emph{factorized spatio-temporal convolutional networks} ($\FSTCN$). A schematic diagram of $\FSTCN$ is shown in Figure~\ref{fig:FSTCN}. While details of $\FSTCN$ will be presented in the next section, we summarize the key characteristics and main contributions of $\FSTCN$ as follows.

\begin{itemize}

\item $\FSTCN$ factorizes the original 3D spatio-temporal convolution kernel learning as a sequential process of learning 2D spatial kernels in the lower network layers, called spatial convolutional layers (SCL), followed by learning 1D temporal kernels in the upper network layers, called temporal convolutional layers (TCL). This factorized scheme greatly reduces the number of network parameters to be learned, thus mitigating the compound difficulty of high kernel complexity and insufficient training video data.

\item We introduce a novel transformation and permutation (T-P) operator to form an intermediate layer of $\FSTCN$, as illustrated by the yellow box in Figure~\ref{fig:FSTCN}. The \mbox{T-P} operator facilitates learning of the temporal convolution kernels in the subsequent TCL.

\item To address the issue of sequence alignment, we propose a training and inference strategy based on sampling multiple video clips from a given action video sequence. Each video clip is produced by temporally sampling with a stride and spatially cropping from the same location of the given action video sequence. Using sampled video clips as inputs to $\FSTCN$ improves the robustness of $\FSTCN$ against variations caused by sequence misalignment.  This is similar in spirit to the data augmentation scheme commonly used for image classification \cite{AlexNet}.
%(cf. Section \ref{Sec-VideoClipSampling})

\item In addition, we propose a novel score fusion scheme based on the sparsity concentration index~(SCI).  It puts more weights on the score vectors of class probability (output of $\FSTCN$) that have higher degrees of sparsity.
%(cf. Section \ref{Sec-ScoreFusion})
Experiments show that this score fusion scheme consistently improves over existing ones.

\end{itemize}

In summary, $\FSTCN$ is a cascaded deep architecture stacking multiple lower SCLs, a T-P operator, and an upper TCL. An additional SCL is also used in parallel with the TCL, aiming at learning a more abstract feature representation of spatial appearance. With the fully-connected (FC) and classifier layers on top, the whole $\FSTCN$ can be trained globally using back-propagation \cite{LeCun98BackProp}. Extensive experiments on benchmark human action recognition datasets \cite{UCF-101,HMDB-51,KTH} show the efficacy of $\FSTCN$.

\section{The proposed deep architecture} \label{Sec-FSTCN}
Human actions in video sequences are 3D signals comprising visual appearance that dynamically evolves over time. To differentiate between different action categories, discriminative spatio-temporal 3D filters are learned to characterize different action patterns. To extend the conventional CNN \cite{lecun} to the spatio-temporal domain, it is necessary to learn a hierarchy of 3D convolution kernels and convolve the learned kernels with the input video data. Concretely, convolving a video cube $\mathbf{V} \in \mathbb{R}^{m_x\times m_y\times m_t}$ with a 3D kernel $\mathbf{K} \in \mathbb{R}^{n_x\times n_y\times n_t}$ can be written as
\begin{equation}\label{Eqn-3DConv}
\mathbf{F}_{st} = \mathbf{V} * \mathbf{K} ,
\end{equation}
where $*$ denotes 3D convolution and $\mathbf{F}_{st}$ is the resulting spatio-temporal features. Ideally a learned kernel $\mathbf{K}$ encodes some primitive spatio-temporal action patterns such that the entire set of action patterns can be reconstructed from sufficiently many such kernels.  However, as discussed in Section \ref{Sec-Intro}, learning a representative set of 3D spatio-temporal convolution kernels is practically challenging due to the compound difficulty of the high complexity of 3D kernels and insufficient training videos. This is in contrast with the problem of learning 2D spatial kernels for image classification \cite{AlexNet}. To overcome this challenge, we resort to approximating the 3D kernels by characterizing the spatio-temporal action patterns using lower-complexity kernels. Not only does this approach need less videos for training, but it can also take advantage of existing massive image datasets to train the spatial kernels.

Although equivalence does not hold in general, we exploit the computational advantage by restricting to a family of 3D kernels $\mathbf{K}$ which can be expressed in a factorized form as
\begin{equation}\label{Eqn-KernelFac}
\mathbf{K} = \mathbf{K}_{x, y} \otimes \mathbf{k}_t ,
\end{equation}
where $\otimes$ denotes the Kronecker product, $\mathbf{K}_{x, y} \in \mathbb{R}^{n_x\times n_y}$ is a 2D spatial kernel and $\mathbf{k}_t \in \mathbb{R}^{n_t}$ is a 1D temporal kernel. Given this kernel factorization, 3D convolution in (\ref{Eqn-3DConv}) can be \emph{equivalently} written as two sequential steps:
\begin{eqnarray}\label{Eqn-Factorized3DConv}
\mathbf{F}_s(:,:,i_t) =  \mathbf{V}(:, :, i_t) * \mathbf{K}_{x, y} , \ i_t = 1, \dots, m_t, \nonumber \\
\mathbf{F}_{st}(i_x, i_y, :) = \mathbf{F}_s(i_x, i_y, :) * \mathbf{k}_t , \ i_x = 1, \dots, m_x, \nonumber \\  i_y = 1, \dots, m_y ,
\end{eqnarray}
where $\mathbf{V}(:, :, i_t)$ denotes an individual frame of $\mathbf{V}$, $\mathbf{F}_s \in \mathbb{R}^{m_x\times m_y\times m_t}$ is obtained by convolving each frame $\mathbf{V}(:, :, i_t)$ with the 2D spatial kernel $\mathbf{K}_{x, y}$ (padding the boundaries of $\mathbf{V}$ before convolution), $\mathbf{F}_s(i_x, i_y, :)$ denotes a vector of $\mathbf{F}_s$ along the temporal dimension, and $\mathbf{F}_{st} \in \mathbb{R}^{m_x\times m_y\times m_t}$ is obtained by convolving each vector $\mathbf{F}_s(i_x, i_y, :)$ with the 1D temporal kernel $\mathbf{k}_t$ (padding the boundaries of $\mathbf{F}_s$ before convolution). Equation (\ref{Eqn-Factorized3DConv}) suggests that one may separately learn a 2D spatial kernel and a 1D temporal kernel and apply the learned kernels sequentially to simulate the 3D convolution procedure. In so doing, the kernel complexity is reduced by an order of magnitude from $n_x n_y n_t$ to $n_x n_y + n_t$, making the learning of effective kernels for action recognition computationally more feasible.
What is more, such a factorized scheme enables the learning of 2D spatial kernels to benefit from the existing massive image datasets \cite{imagenet} with which the performance of many vision tasks \cite{girshick2014rcnn, dl-segmentation, dl-retrieval} have been boosted significantly. We note that since the ranks of the 3D kernels constructed by (\ref{Eqn-KernelFac}) are generally lower than those of the general 3D kernels, it appears that we sacrifice the representation power by using the factorized scheme.  However, spatio-temporal action patterns in general have a low-rank nature, since feature representations of static appearance of human actions are largely correlated across nearby video frames. In case that they are not of low-rank, the sacrificed representation power may be compensated by learning redundant 2D and 1D kernels and constructing candidate 3D kernels from them.
\subsection{The details of the architecture}
The above analysis motivates us to design a novel deep architecture which factorizes multiple layers of the original 3D spatio-temporal convolutions as a sequential process involving 2D spatial convolutions followed by 1D temporal convolutions.  More specifically, our proposed $\FSTCN$ consists of several spatial convolutional layers (SCL).  The basic components of an SCL are 2D spatial convolution kernels \footnote{Considering multiple feature channels or maps for each video frame, the convolution kernels in SCLs are in fact 3D kernels. To conceptually keep consistent with the 3D physical space, we choose to term the convolution kernels in SCLs as 2D kernels. The same reason applies to terming the convolution kernels in TCLs as 1D kernels.}, nonlinearity (ReLU), local response normalization and max-pooling, as illustrated in the black box of Figure \ref{fig:FSTCN}. Each convolutional layer must include convolution and ReLU but normalization and max-pooling are optional. By processing individual frames of the video clips with the learned 2D spatial kernels, the SCLs are able to extract compact and discriminative features for the visual appearance.

To characterize the motion patterns, $\FSTCN$ further stacks a temporal convolutional layer (TCL) on top of the SCLs. The TCL has the same basic components as those of the SCLs. In order to learn the motion patterns that evolve over time, a layer called the T-P operator is inserted between the SCLs and TCL, as illustrated in the yellow box of Figure \ref{fig:FSTCN}. Taking data in the form of 4D arrays (horizontal $x$, vertical $y$, temporal $t$, and feature channel $f$ as dimensions) as input, this T-P operator first vectorizes individual matrices of the 4D arrays along the horizontal and vertical dimensions such that each matrix of size $x \times y$ becomes a vector of length $x\times y$, and then rearranges the dimensions of the resulting 3D arrays (\emph{the transformation operation}) so that 2D convolution kernels can be learned and applied along the temporal and feature-channel dimensions (i.e., the 1D temporal convolution in TCL).\footnote{We term the convolution in TCL as 1D temporal convolution to conceptually keep it consistent with the 3D physical space. This convolution is in fact 2D convolution along the temporal and feature-channel dimensions.}
Note that the transformation operation is optional; our introduction of it in the T-P operator is conceptually to make temporal convolution in the subsequent TCL explicit, and practically to make the implementation of $\FSTCN$ compatible with the popular deep learning libraries \cite{AlexNet,jia2014caffe}. Vectorization and transformation are followed by a permutation operation, via a permutation matrix $P$ with a size of $f\times f^{'}$, along the dimension of feature channels. It aims to reorganize the output of feature channels so that convolution in the subsequent TCL takes a better support of local 2D windows in the feature-channel and temporal directions. As tunable network parameters, $P$ is initialized from Gaussian distribution and learned, in the same way as other network parameters, via back-propagation. Consequently, it reorganizes the feature channels by generating $f^{'}$ new feature maps via weighted combination of the $f$ input feature maps. Since TCL takes as input the output of the T-P operator, which in turn takes as input the intermediate feature maps of all frames of the input video clip (i.e., output of the SCLs), a pixel location in the vectorized spatial domain of TCL corresponds to a larger receptive field of the input video clip. In other words, TCL's 1D temporal convolution kernels essentially feature the motion patterns constituted by the visual appearance of local regions of the input video clip that evolves over time. When combined with our proposed video clip sampling strategy (to be presented in Section \ref{Sec-VideoClipSampling}), they capture long-range motion patterns of relatively holistic visual appearance at a cheaper learning cost. Details of the TCL are presented in the purple box of Figure \ref{fig:FSTCN}. Two parallel convolutions with different kernel sizes are applied to the TCL and then concatenated together to represent the temporal characteristics. Dropout follows each ReLU, respectively, to reduce overfitting. Two advantages can be observed.  First, as stated before, actions can be performed at different speeds or with varying accelerations, that is, the ``slow'' ones (long time duration) can be captured using the large kernel while the ``fast'' ones (short time duration) using the small kernel. What is more, the parallel convolutional layers can provide more motion properties which will definitely benefit the action recognition task.

In $\FSTCN$, an additional SCL is also used in parallel with TCL, aiming at learning a more abstract feature representation of visual appearance. Similar to a convolutional layer in conventional CNNs, this SCL improves the spatial invariance of action categories by extracting salient/discriminative appearance features via the learned spatial filters and the subsequent nonlinearity and pooling operations. Two fully-connected layers are stacked on top of the parallel TCL and SCL, which are then concatenated as the spatio-temporal features. Finally, a FC layer and a softmax classification layer are further cascaded for supervised training by standard back-propagation. The whole architecture of $\FSTCN$ with specific layer components is presented in Figure \ref{fig:FSTCN}.

\subsection{Data augmentation by sampling video clips} \label{Sec-VideoClipSampling}

Human actions are visual signals contained in video sequences. Some of them are also periodic with multiple action cycles repeating over time. It is usually a pre-requisite step to detect and align action instances from the containing video sequences, in order to compare action instances performed in different speeds or with varying accelerations. This issue of action sequence detection/alignment is traditionally addressed by sliding windows across the temporal direction, dynamic time warping \cite{DTW}, or detecting trajectories of interest points from video sequences \cite{st_interest_points,improved-trajectories}. However, it is generally overlooked in the more recent deep learning based action recognition methods \cite{3dCNN,vid_cnn,TwoStream}.

\begin{figure}
  \centering
  \centerline{\includegraphics[width=8.5cm]{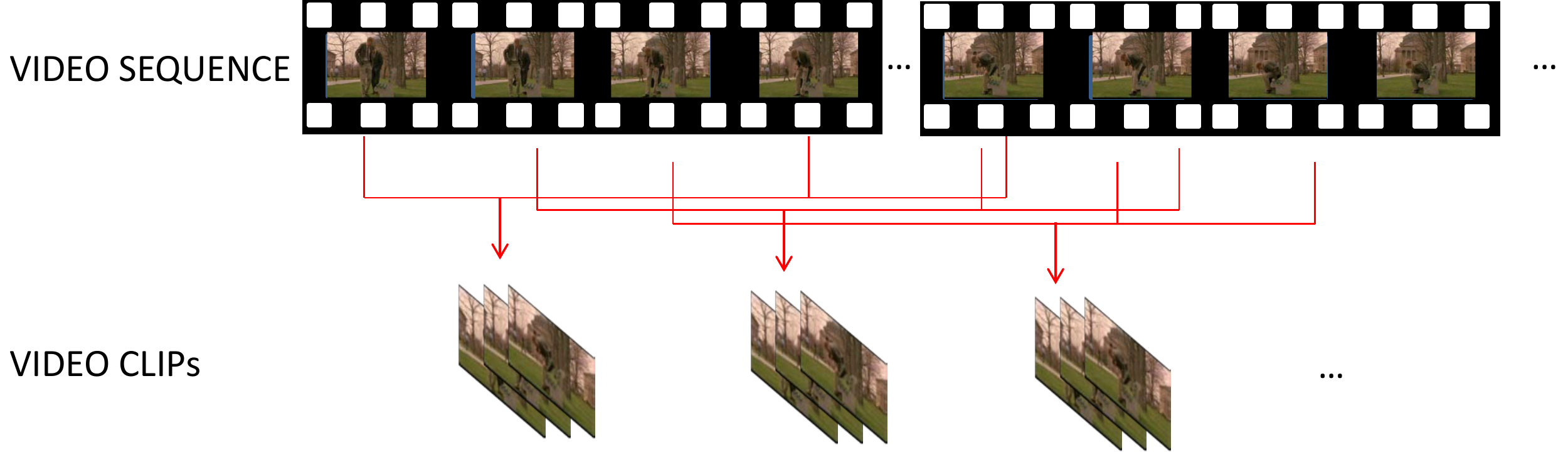}}
  \caption{Illustration of our proposed video clip sampling scheme. Each video clip is produced by \emph{temporally sampling with a
stride} and \emph{spatially cropping} from the same location of the given video sequence. Numbers of pairs of video clips $\{ \mathbf{V}_{clip}, \mathbf{V}_{clip}^{diff} \}$ consist our $\FSTCN$ input.}\label{fig:clips}\vspace{-1.5em}
\end{figure}

Instead of deliberately detecting and aligning action instances from video sequences, we propose in this paper a training and inference strategy based on sampling multiple video clips from a given video sequence. Note that our proposed scheme is different from the bag of video words \cite{Niebles08}, which extracts spatio-temporal features from the \emph{video cuboid}. Each video clip in our scheme is produced by \emph{temporally sampling with a stride} and \emph{spatially cropping} from the same location of the given video sequence, as illustrated in Figure \ref{fig:clips}. Such sampled video clips are not guaranteed to be aligned with action cycles. However, the motion pattern is well kept in the sampled video clips if enough time duration is given. We use them to train $\FSTCN$ in a supervised manner, and expect representations robust to the misalignment can be learned at the upper network layers. Besides, even some misalignment exists, since our TCL learns the kernel along the feature and temporal dimensions, the discriminative motion patterns can still be preserved in series.

More specifically, given a video sequence $\mathbf{V} \in \mathbb{R}^{m_x\times m_y\times m_t}$, a video clip $\mathbf{V}_{clip} \in \mathbb{R}^{l_x\times l_y\times l_t}$ ($l_x < m_x, l_y < m_y$, and $l_t < m_t$) is randomly sampled from $\mathbf{V}$ by spatially specifying a patch of size $l_x\times l_y$ and temporally specifying a starting frame, and regularly cropping $l_t$ frames of $\mathbf{V}$ with a temporal stride of $s_t$.  Such a sampled $\mathbf{V}_{clip}$ approximates human actions contained in a 3D cube of size $l_x\times l_y\times (l_t - 1)s_t$. However, it only conveys long-range motion dynamics when the temporal stride $s_t$ is relatively large. To add in the information of short-range motion dynamics, for each frame $\mathbf{V}(:, :, i_t)$ of $\mathbf{V}$ with $i_t \in \{1, \dots, m_t \}$, we also compute
\begin{equation}\label{Eqn-FrameDiff}
\mathbf{V}^{diff}(:, :, i_t) = \left| \mathbf{V}(:, :, i_t) - \mathbf{V}(:, :, i_t + d_t) \right| ,
\end{equation}
where $\mathbf{V}^{diff}$ denotes the resulting sequence and $d_t$ is the distance of the two frames to compute the difference. We sample from $\mathbf{V}^{diff}$ a video clip $\mathbf{V}_{clip}^{diff} \in \mathbb{R}^{l_x\times l_y\times l_t}$, in the same way as sampling $\mathbf{V}_{clip}$ from $\mathbf{V}$ and with a same sampling index set of $\left\{ i_t \in \{1, \dots, m_t\} \right\}$. Note that $\mathbf{V}_{clip}^{diff}$ conveys both short-range and long-range motion information \footnote{Individual frames of $\mathbf{V}_{clip}^{diff} \in \mathbb{R}^{l_x\times l_y\times l_t}$ convey short-range motion information, while $\mathbf{V}_{clip}^{diff}$ as a whole conveys long-range motion by covering an extended duration (size $(l_t - 1)s_t$) of video sequence.}. An illustration of our video clip sampling scheme is presented in Figure \ref{fig:clips}.

We sample multiple pairs of video clips $\{ \mathbf{V}_{clip}, \mathbf{V}_{clip}^{diff} \}$, and use the sampled video clip pairs as inputs of $\FSTCN$. Our sampling strategy is spiritually similar to data augmentation commonly used in image classification \cite{AlexNet}, where results have demonstrated that such a strategy is able to reduce overfitting of network parameters by largely increasing training samples, and also to improve robustness of the learned networks against misalignment of object instances in images. Our proposed video clip sampling strategy extends data augmentation to the temporal domain, aiming to address the issue of sequence alignment in the problem of video based human action recognition.

Considering that $\mathbf{V}_{clip}^{diff}$ contains short-range and long-range motion information, and $\mathbf{V}_{clip}$ (mostly) contains information of visual appearance, our use of the sampled video clip pairs in $\FSTCN$ is as follows. We first feed individual frames of each pair of $\mathbf{V}_{clip}$ and $\mathbf{V}_{clip}^{diff}$ into the lower SCLs. After mid-level spatial feature representations of these individual frames are extracted, we separate these mid-level features by feeding those from all frames of $\mathbf{V}_{clip}^{diff}$ into TCL (after T-P operator), and feeding a randomly sampled frame of $\mathbf{V}_{clip}$ into the intermediate SCL that is parallel to the TCL. When testing, the selected middle one of $\mathbf{V}_{clip}$ is fed into SCL. This separation of signal pipelines is consistent with our architecture design of $\FSTCN$.

\subsection{Learning and inference}

$\FSTCN$ has factorized SCLs and TCLs. To learn spatial and temporal convolution kernels effectively, we follow the ideas from \cite{GoogLeNet} and introduce auxiliary classifier layers connected to the lower SCLs, as illustrated in the dashed red box of Figure \ref{fig:FSTCN}. In practice, we first use ImageNet \cite{imagenet} to pre-train this auxiliary network, and use randomly sampled training video frames to fine-tune it, in order to get better 2D spatial convolution kernels in the lower SCLs. We follow the advice in \cite{vid_cnn} by only fine-tuning the last three layers of the auxiliary network as well. Finally the whole $\FSTCN$ network is \emph{globally} trained via back-propagation, using sampled pairs of video clips as inputs. Note that in our training of TCL in $\FSTCN$, we do not use video data additional to the currently working action dataset; in contrast, additional training videos from a second action dataset are used for training in two-stream CNNs \cite{TwoStream}.

In the inference stage, given a test action sequence, we first sample pairs of video clips as explained in Section \ref{Sec-VideoClipSampling}. We then pass each of the sampled video clip pairs through the $\FSTCN$ pipeline, resulting in a score output of class probability. These scores are fused to get the final recognition result, for which we propose a novel score fusion scheme that will be introduced shortly.

\subsection{SCI based score fusion}\label{Sec-ScoreFusion}

\begin{figure}
  \centering
  \centerline{\includegraphics[width=8.5cm]{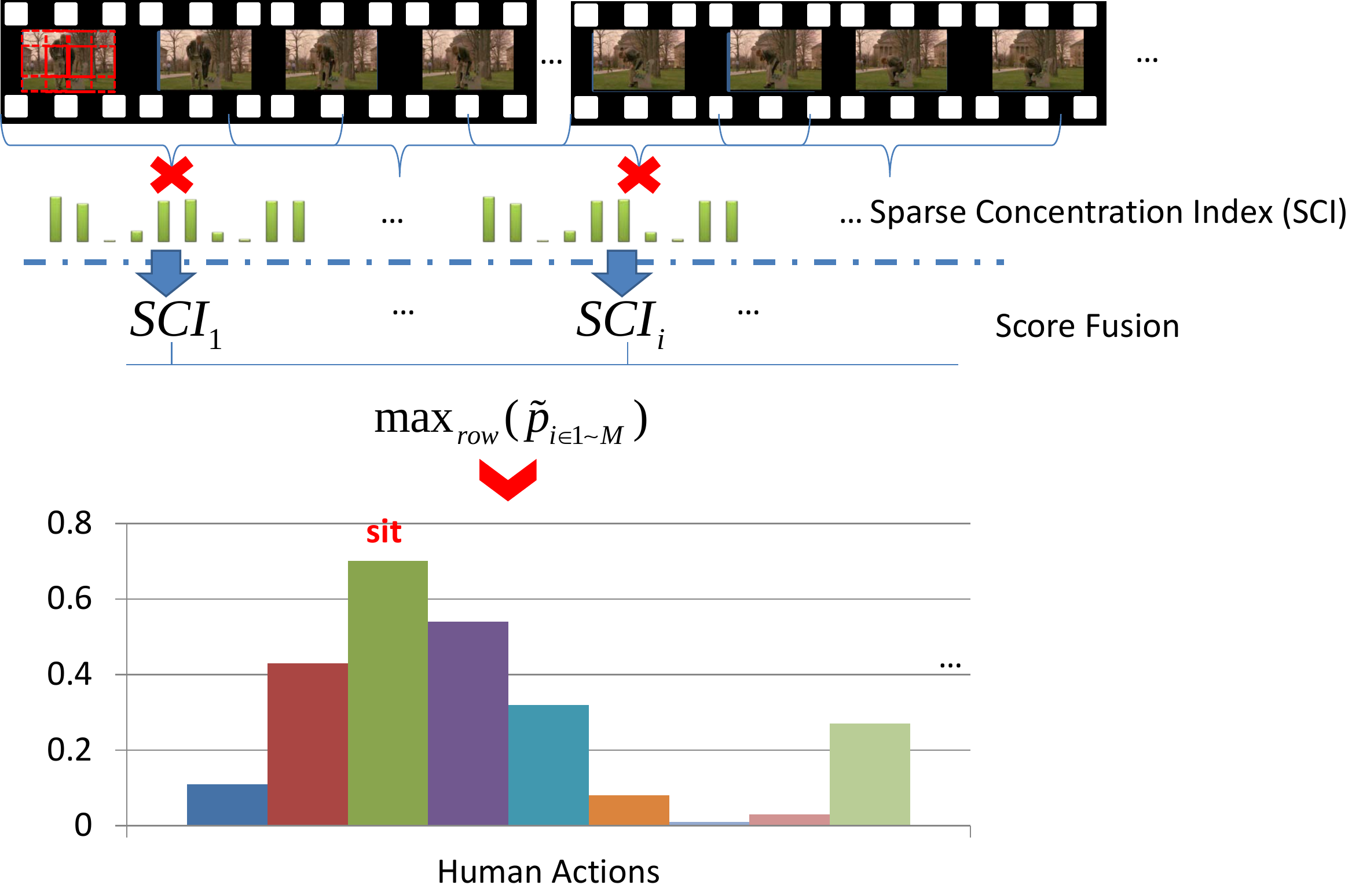}}
  \caption{The SCI based score fusion scheme. In the stage of inference, given a test action sequence, we first sample pairs of video clips of the video sequence. Each pair of video clip is cropped from top left, top middle, top right, middle left, middle, middle right, bottom left, bottom middle, bottom right, forming 9 parts and flipped to generate 18 samples which will pass through the $\FSTCN$ pipeline, resulting in 18 scores output of class probability fused using SCI. All the output scores from the sampled pairs are then maximized to generate the final score of class probability $\mathbf{p} \in \mathbb{R}^N$. The action category corresponds to the largest entry value of $\mathbf{p}$.}\label{fig:SCI}\vspace{-1.5em}
\end{figure}

Assume there are $N$ categories of human actions in an action recognition dataset, and we sample $M$ pairs $\{ \mathbf{V}_{clip}, \mathbf{V}_{clip}^{diff} \}$ of video clips from each video sequence of the dataset. Each pair of video clips is regularly cropped (one sampling strategy is presented in Figure \ref{fig:SCI}), generating $C$ crops similarly as \cite{AlexNet} on ImageNet. For a test video sequence, denote the score output of class probability for the $k^{th}$ cropped video of the $i^{th}$ sampled video clip pair as $\mathbf{p}_{k,i} \in \mathbb{R}^N$, with $k \in \{1, \dots, C\}$ and $i \in \{1, \dots, M\}$. The final score $\hat{\mathbf{p}}$ of class probability can be obtained by simple averaging, i.e., $\hat{\mathbf{p}} = \frac{1}{CM}\sum_{i=1}^M \sum_{k=1}^C\mathbf{p}_{k, i}$. This scheme presumes that contributions from each of $\{ \mathbf{p}_{k,i} \}{_{k=1}^C}{_{i=1}^M}$ are of equal importance, which is however, not usually true. Indeed, if one has knowledge about which score outputs of $\{ \mathbf{p}_{i} \}{_{k=1}^C}{_{i=1}^M}$ are more reliable, a weighted averaging scheme can be developed to get a better final score $\hat{\mathbf{p}}$.

To estimate the degree of reliability for any score $\mathbf{p} \in \mathbb{R}^N$ of class probability, we come up with a very intuitive idea: when $\mathbf{p}$ is reliable, it is usually sparse with low entropy of the distribution, i.e., only a few entries  of the vector $\mathbf{p}$ have large values (meaning the probabilities that the test video sequence is of the corresponding action categories are high), while values of other entries in $\mathbf{p}$ are small or approaching zeros; conversely, when $\mathbf{p}$ is not reliable, its entry values (class probabilities) tend to spread evenly over all the action categories. This presumption suggests that we may use the sparsity degrees of each of $\{ \mathbf{p}_{k,i} \}{_{k=1}^C}{_{i=1}^M}$ to derive a weighted score fusion scheme. To this end, we introduce the notion of Sparsity Concentration Index (SCI) \cite{SCIInRobustFaceRecog} to measure the degree of sparsity for any $\mathbf{p}$, which computes,

\begin{equation}\label{Eqn-SCI}
\textrm{SCI}(\mathbf{p}) = \frac{N\cdot \max_{j = 1, \dots, N } p_{j} / \sum_{j=1}^N p_{j} - 1}{N - 1} ,
\end{equation}
where $p_{j}$ denotes the $j^{th}$ entry of $\mathbf{p}$, and $\textrm{SCI}(\mathbf{p}) \in [0, 1]$. Given $C$ cropped videos of the $i^{th}$ sampled video clip pair and their score outputs $\{ \mathbf{p}_{k,i} \}_{k=1}^C$, our proposed SCI based score fusion scheme computes the final score of class probability as
\begin{equation}\label{Eqn-SCIScoreFusion-1}
\mathbf{p_i} = \frac{ \sum_{k=1}^C\textrm{SCI}(\mathbf{p}_{k,i}) \mathbf{p}_{k,i} }{ \sum_{k=1}^C\textrm{SCI}(\mathbf{p}_{k,i}) } .
\end{equation}
The $M$ pairs of video clips are fused by
\begin{equation}\label{Eqn-SCIScoreFusion-2}
\hat{\mathbf{p}} = \mathrm{max}_{\mathrm{row}}( [\tilde{\mathbf{p}}_{1}, \dots, \tilde{\mathbf{p}}_{M}] ) ,
\end{equation}
The test video sequence is finally recognized as the action category that has the corresponding largest entry value of $\hat{\mathbf{p}}$, i.e., $\arg\max_{j=1, \dots, N}\hat{p}_j$ with $\hat{p}_j$ as the $j^{th}$ entry of $\hat{\mathbf{p}}$. Our score fusion scheme also provides the compensation of the misalignment problem since maximized values of video clips are taken.

We illustrate in Figure \ref{fig:SCI} the idea of our proposed SCI based score fusion scheme. Experiments in Section \ref{Sec-Exps} show that it consistently improves over the commonly used averaging scheme. We expect our proposed scheme is also useful in other deep learning based classification methods.

\subsection{Implementation details}
\label{Sec-ImplementDetail}
The details of the first four SCLs which extract the compact and discriminative feature representations of visual appearance are: $Conv(96,7,2)-ReLU-Norm-Pooling(3,2)-Conv(256,5,2)-ReLU-Norm-Pooling(3,2)-Conv(512,3,1)-Conv(512,3,1)$, where $Conv(c_f,c_{k}, c_{s})$ denotes the a convolutional layer with $c_f$ feature maps and the kernel size is $c_{k}\times c_{k}$, applied to the input with the stride $c_{s}$ in width and height direction. $Norm$ and $Pooling(p_k,p_s)$ are the local response normalization layer and pooling layer in which $p_k$ is the spatial window and $p_s$ is the pooling stride, respectively. Similar as \cite{AlexNet} rectified activation functions (ReLU) are applied to all the hidden weights layers; max-pooling is performed over $3\times3$ spatial windows with stride 2; local response normalization across channels uses the same settings as \cite{AlexNet}, that is: $k=2, n=5, \alpha = 5\times10^{-4}, \beta = 0.75$. The SCL which connects to the TCL contains convolution ($Conv(128,3,1)$) and pooling ($Pooling(3,3)$). The permutation matrix $P$ has the size of $128\times 128$, that is, $f=f^{'}=128$. The TCL has two parallel convolutional layers ($Conv(32,3,1)$ and $Conv(32,5,1)$), each of which has a dropout layer with a dropout probability of 0.5. Note that the TCL will not be followed by pooling layer since pooling will ruin the temporal cues. Two fully-connected (FC) layers with 4096 and 2048 hidden nodes respectively are stacked on top of the TCL and SCL. The feature outputs from the fully connection layer of SCL and TCL are then concatenated and passed through another FC layer with 2048 hidden nodes. For training, we use a batch size of 32, momentum of 0.9, and weight decay of 0.0005. Instead of using the popular input size of $224\times224$, we use the size of $204\times204$ to save memory. Note that the settings of the spatial convolutional path are the same as in \cite{TwoStream} except that the input size becomes smaller. At each training iteration, the frames in each pair of video clips are randomly cropped at the same location and flipped simultaneously.

\section{Experiments}
\label{Sec-Exps}

Experiments are conducted on two benchmark action recognition datasets, namely, UCF-101 \cite{UCF-101} and HMDB-51 \cite{HMDB-51}, which are the largest and most challenging annotated action recognition datasets.

\textbf{UCF-101 } \cite{UCF-101} is composed of realistic web videos, which are typically captured with camera motions and under various illuminations, and contain partial occlusion. It has 101 categories of human actions, ranging from daily life to unusual sports (such as ``Yo Yo''). UCF-101 has more than 13K videos with an average length of 180 frames per video. It has 3 split settings to separate the dataset into training and testing videos. We report the mean classification accuracy over these three splits.

\textbf{HMDB-51 } \cite{HMDB-51} has a total of 6766 videos organized as 51 distinct action categories, which are collected from a wide range of sources. This dataset is more challenging than others because it has more complex backgrounds and context environments. What is more, there are many similar scenes in different categories. Since the number of training videos is small in this dataset, it is more challenging to learn representative features. Similar to UCF-101, HMDB-51 also has 3 split settings to separate the dataset into training and testing videos, and we report the mean classification accuracy over these three splits.

\begin{table}[htb] \caption{Results of TCL path of $\FSTCN$ and optical flow stream CNN of \cite{TwoStream} on HMDB-51 (split 1)} \label{t:com-kernel}
\footnotesize
\centering \vspace{0.3cm}
\begin{tabular}{c|c} \hline
Training setting & Accuracy     \\
\hline
TCL (only $3\times3$ kernel)                        & 46.0\% \\
TCL (only $5\times5$ kernel)                        & 47.1\% \\
TCL ($5\times5$ kernel + $3\times3$ kernel)         & 48.4\% \\
Training on HMDB-51 without additional data \cite{TwoStream}          & 46.6\% \\
Fine-tuning a ConvNet, pre-trained on UCF-101 \cite{TwoStream}        & 49.0\% \\
Training on HMDB-51 with classes added from UCF-101 \cite{TwoStream}  & 52.8\% \\
Multi-task learning on HMDB-51 and UCF-101 \cite{TwoStream}           & 55.4\% \\
\hline
\end{tabular}
\vspace{-1.5em}
\end{table}
In the experiments, the video clip consists of $5$ temporally sampled pairs of video clips $\{ \mathbf{V}_{clip}, \mathbf{V}_{clip}^{diff} \}$ with $d_t = 9$ and $s_t = 5$. These sampled video clips are representative enough to convey the long-range motion dynamics. Firstly, we use the TCL path of $\FSTCN$ (orange arrows in Figure \ref{fig:FSTCN}) and split 1 of HMDB51 to investigate whether using two different convolution kernels in TCL is better than using one kernel only. The sizes of the two different kernels are $3\times 3$ and $5\times 5$ respectively. Results of this investigation are reported in Table \ref{t:com-kernel}. Table \ref{t:com-kernel} tells that using two different kernels is better than using either one of them, and using kernel of a larger size is better than using that of a smaller size. Note that these results are obtained without using our score fusion scheme. In Table \ref{t:com-kernel}, we also compare with \cite{TwoStream} in the setting of using temporal convolution pipeline only, i.e., using the TCL path with $\mathbf{V}_{clip}^{diff}$ as the input for $\FSTCN$ and the optical flow CNN stream for \cite{TwoStream}. Our result ($48.4\%$) of TCL path is better than that ($46.6\%$) of optical flow CNN stream in \cite{TwoStream} when only split 1 of HMDB-51 is used as training videos. We note that auxiliary training videos are also used in \cite{TwoStream} to boost performance, as shown in Table \ref{t:com-kernel}. Using auxiliary training videos is complementary to our proposed technique of factorized SCL and TCL. We expect our result can be further improved given auxiliary training videos.

\begin{table}[htb] \caption{Mean accuracy on UCF-101 and HMDB-51 using different strategies of $\FSTCN$} \label{t:SCL-TCL}
\footnotesize
\centering \vspace{0.3cm}
\begin{tabular}{c|c|c} \hline
Training setting & UCF-101 & HMDB-51    \\
\hline
only SCL path                                  & 71.3\% & 42.0\% \\
only TCL path                                  & 72.6\% & 45.8\% \\
only TCL path (SCI fusion)                     & 76.0\% & 49.3\% \\
$\FSTCN$ (single randomly selected clip)       & 84.5\% & 54.1\% \\
$\FSTCN$ (averaging fusion)                    & 87.9\%  & 58.6\% \\
$\FSTCN$ (SCI fusion)                          & 88.1\%  & 59.1\% \\
\hline
\end{tabular}
\end{table}
We present controlled experiments on the UCF-101 and HMDB-51 datasets in Table \ref{t:SCL-TCL}, where results from different proposed contributions are specified. Table \ref{t:SCL-TCL} tells that our proposed data augmentation scheme by sampling video clips, and also the SCI based score fusion scheme effectively improve the recognition performance. In particular, when features from SCL path and TCL path are concatenated and trained globally via back-propagation, about $10\%$ gain can be obtained, indicating that our learned spatio-temporal features are complementary with each other. Results from our main contribution of $\FSTCN$ will be presented shortly by comparing with the state-of-the-art.

Table \ref{t:CompResultTable} compares $\FSTCN$ with other state-of-the-art methods, where performance is measured by mean accuracy on three splits of the HMDB51 and UCF101 datasets. Compared with the state-of-the-art CNN based method \cite{TwoStream}, our method outperforms it by about $1\%$ on both datasets, when averaging fusion is adopted. When a supervised learning based SVM score fusion scheme is used in \cite{TwoStream}, our method still achieves better or comparable performance on the two datasets. We note that the results of \cite{TwoStream} and \cite{vid_cnn} are obtained by using auxiliary training videos, while our results are obtained by using each of the working datasets only. We expect our results can be further boosted given auxiliary training videos.

%\begin{table}[htb] \caption{Classification accuracy on HMDB-51 (split 1) using TCL with different kernels} \label{t:com-kernel}
%\small
%\centering \vspace{0.3cm}
%\begin{tabular}{c|c} \hline
%Training setting & Accuracy     \\
%\hline
%TCL (only $5\times5$ kernel)                      & 47.1\% \\
%TCL (only $3\times3$ kernel)                      & 46.0\% \\
%TCL ($5\times5$ kernel + $3\times3$ kernel)       & 48.4\% \\
%\hline
%\end{tabular}
%\end{table}

\begin{table}[htb] \caption{Classification mean accuracy (over three splits) on UCF-101 and HMDB-51.} \label{t:CompResultTable}
\footnotesize
\centering \vspace{0.3cm}
\begin{threeparttable}
\begin{tabular}{c|c|c} \hline
Methods & UCF-101 & HMDB-51 \\
\hline
Improved dense trajectories (IDT) \cite{improved-trajectories}   & 85.9\%    & 57.2\% \\
IDT higher-dimensional encodings  \cite{high-encodings}         & 87.9\%    & 61.1\% \\
Spatio-temporal HMAX network \cite{HMAX} \cite{HMDB-51}          & -\%       & 22.8\% \\
"Slow fusion" spatio-temporal ConvNet \cite{vid_cnn}             & 65.4\%    & -\%    \\
Two-stream model (averaging fusion) \cite{TwoStream}             & 86.9\%    & 58.0\% \\
Two-stream model (SVM fusion) \cite{TwoStream} \tnote{*}         & 88.0\%    & 59.4\% \\
$\FSTCN$ (averaging fusion)                                     & \textbf{87.9}\%  & \textbf{58.6}\% \\
$\FSTCN$ (SCI fusion)                                           & \textbf{88.1}\%  & \textbf{59.1}\% \\
\hline
\end{tabular}
\begin{tablenotes}
\scriptsize
\item[*] Additional videos are fed into the network to train the optical flow stream since multi-task learning strategy is applied.
\end{tablenotes}
\end{threeparttable}
\end{table}
\vspace{-1.5em}

\section{Visualization}

\begin{figure}
  \centering
  \centerline{\includegraphics[width=8cm]{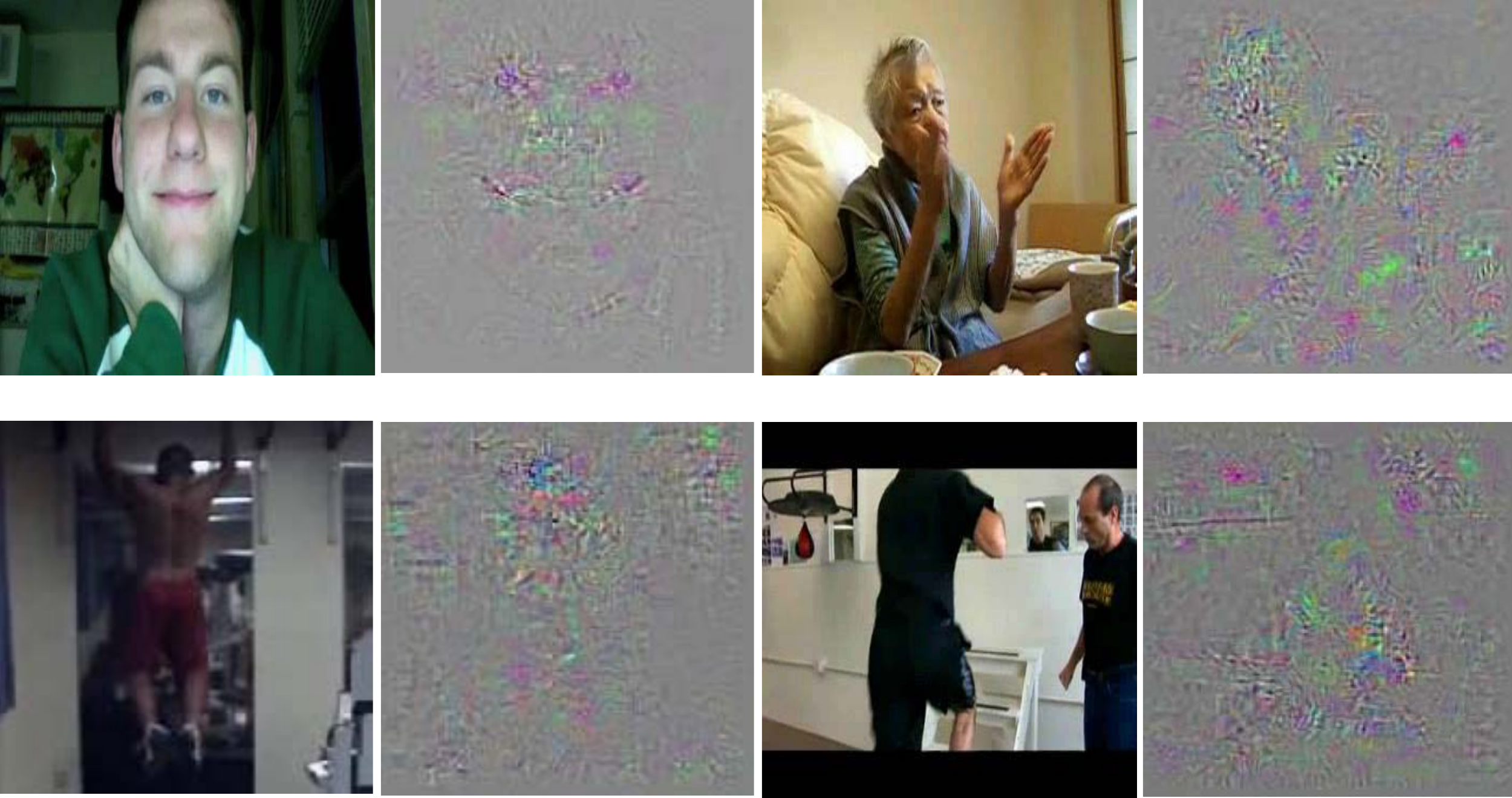}}
  \caption{The visualization of the ``saliency'' map which maximize the output score, from left to right, from top to down, the category is smile, clap, pull-up and climbing. Better to be seen in the color version.}\label{fig:saliencymap}\vspace{-1.5em}
\end{figure}

To visually verify the relevance of learned parameters in $\FSTCN$, we use back-propagation to visualize important regions for any action category, i.e., back-propagating the neuron of that action category in the classifier layer to the input image domain. Figure \ref{fig:saliencymap} gives illustration for several action categories. The shown ``saliency'' maps suggest that learned parameters in $\FSTCN$ can capture the most representative regions of action categories. For example, the saliency map of the action ``smile'' displays a ``monster'' face, suggesting that ``smile'' happens around the mouth.

\begin{figure}
  \centering
  \centerline{\includegraphics[width=8cm]{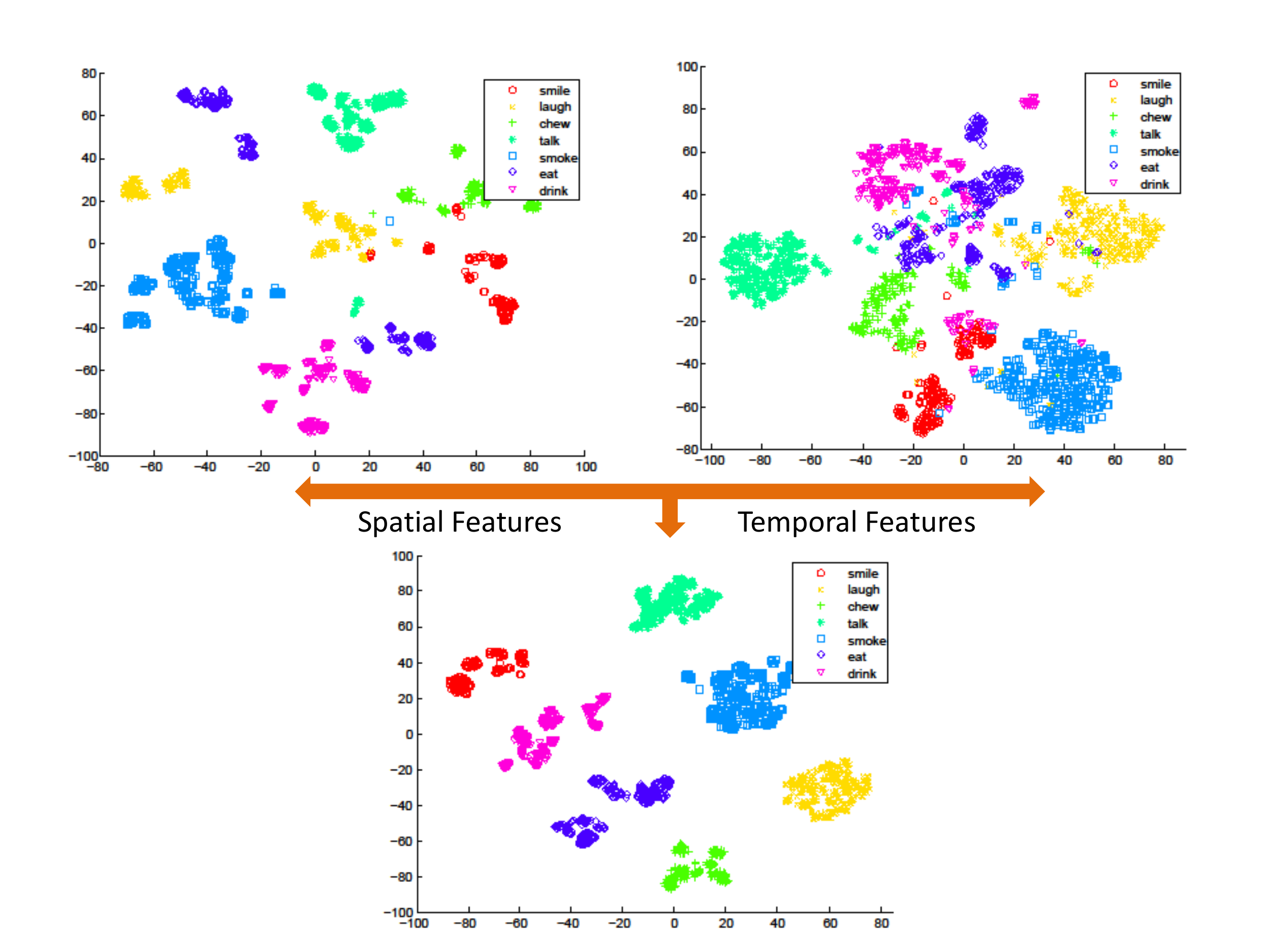}}
  \caption{The feature visualization of seven categories from HMDB-51. These seven categories focus on the tiny mouth movement and they have the same scene with the big head in the center. Spatial features are extracted from the second FC layer of SCL and temporal features are from the second FC layer of TCL, and spatio-temporal features are from the FC layer after SCL and TCL being concatenated. All the features are visualized using the state-of-the-art method tSNE which can be viewed better in color.}\label{fig:vi-features-7}\vspace{-0.5em}
\end{figure}

To investigate whether our learned spatio-temporal features are discriminative for action recognition, we plot in Figure \ref{fig:vi-features-7} the learned features of several action categories (``smile'', ``laugh'', ``chew'', ``talk'', ``eat'' , ``smoke'', and ``drink'' in the HMDB-51 dataset). These features are visualized using the dimensionality reduction method tSNE \cite{tSNE}. Since these action categories are mainly concerned with face motions, especially with mouth movements, they cannot be easily distinguished. Figure \ref{fig:vi-features-7} clearly shows that spatio-temporal features extracted from the FC layer after SCL and TCL being concatenated, are more discriminative than either spatial features extracted from the second FC layer of SCL, or temporal features extracted from the second FC layer of TCL.

%\begin{figure}
%  \centering
%  \centerline{\includegraphics[width=7cm]{figures/conv1_01.pdf}}
%  \caption{The learned temporal filter with large kernel size, in order to make it clear the figure is resized using the factor of 2.}\label{fig:tfilter1}
%\end{figure}

%\begin{figure}
%  \centering
%  \centerline{\includegraphics[width=7cm]{figures/conv1_02.pdf}}
%  \caption{The learned first-layer temporal filter with small kernel size, in order to make it clear the figure is resized using the factor of 2.}\label{fig:tfilter2}
%\end{figure}

%-------------------------------------------------------------------------
%\begin{figure*}
%  \centering
%  \centerline{\includegraphics[width=10cm]{figures/CM.pdf}}
%  \caption{Confusion Matrix of the proposed method on the first split of HMDB-51.}\label{fig:CM}
%\end{figure*}

\section{Conclusion}
In this paper, a novel deep learning architecture, termed $\FSTCN$, is proposed for action recognition. The $\FSTCN$ is a cascaded deep architecture which learns the effective spatio-temporal features through training using standard back-propagation. This factorization design mitigates the compound difficulty of high kernel complexity and insufficient training videos. The T-P operator provides a novel feature and temporal representation for actions. Moreover, two parallel kernels in the TCL assists it to learn more representative temporal features. In addition, the additional SCL extracts more abstract spatial appearance which largely compensates the deficiency of TCL as shown in the experimental results. Extensive experiments on the action benchmark datasets present the superiority of our algorithm even without additional training videos.
\section{Acknowledgment}
This research has been partially supported by Faculty Research Award Z0400-D granted to Dit-Yan Yeung and the National Natural Science Foundation of China (Grant No. 61202158).

{\small
\bibliographystyle{ieee}
\bibliography{FstCN}
}

\end{document}